\documentclass[letterpaper, 10 pt, conference]{ieeeconf}  % Comment this line out if you need a4paper

\usepackage{times}
\usepackage{amsmath}

\usepackage{array}
\usepackage{multicol}
\usepackage{hyperref}
\usepackage{wrapfig}
\usepackage{amssymb}
\usepackage{xcolor}

\usepackage{tabularx}
\usepackage{acro}
\usepackage{algorithm}
\usepackage{multirow}
\usepackage[capitalise, nameinlink]{cleveref}
\usepackage{cuted}      % For wide figures
\usepackage{graphicx}   % For images
\usepackage{caption}    % For captions
\usepackage{booktabs}
\usepackage{tabularx}
\usepackage{makecell}
\usepackage{tabularx}
\usepackage{subcaption}
\usepackage[dvipsnames,table,xcdraw, HTML]{xcolor}
\usepackage[noend]{algpseudocode} % algorithmic, \State{}, etc
\usepackage{tikz}
\usepackage{etoolbox}
\newcommand*\textcir[1]{\tikz[baseline=(char.base)]{\node[shape=circle,draw,inner sep=0.5pt] (char) {#1};}}
\newcommand{\textccir}[1]{%
  \ifthenelse{\equal{#1}{1}}%
    {\textcolor{black}{\textcir{#1}}}
    {\ifthenelse{\equal{#1}{2}}%
        {\textcolor{black}{\textcir{#1}}}
    {\ifthenelse{\equal{#1}{3}}
    {\textcolor{black}{\textcir{#1}}}
    {\ifthenelse{\equal{#1}{4}}
    {\textcolor{black}{\textcir{#1}}}
    {\ifthenelse{\equal{#1}{5}}
    {\textcolor{black}{\textcir{#1}}}
    {\ifthenelse{\equal{#1}{6}}
    {\textcolor{black}{\textcir{#1}}}
    {#1}}}}}}%
}
\newcommand{\AlgName}{\textsc{NavMoE}}

\usepackage{amsthm}
\usepackage{thmtools}

\declaretheoremstyle[
  headfont=\bfseries,
  headpunct={},
  headindent=0pt,
  spaceabove=6pt,
  spacebelow=6pt,
  notebraces={(}{)},
  headformat=\NAME\ \NUMBER\ \NOTE,
  postheadspace=1em,
  mdframed={skipabove=1em, skipbelow=1em}
]{breaktitle}

\newtheorem{proposition}{Proposition}

\newcolumntype{?}{!{\vrule width 1.2pt}}

\definecolor{anotBlue}{HTML}{001272}
\definecolor{mygreen}{rgb}{0.667, 0.8, 0.6}
\definecolor{myblue}{HTML}{B1D4E0}
\IEEEoverridecommandlockouts  

\algdef{SE}[PROCEDURE]{Procedure}{EndProcedure}%
   [2]{\algorithmicprocedure\ \textproc{#1}\ifthenelse{\equal{#2}{}}{}{(#2)}}%
   {\algorithmicend\ \algorithmicprocedure}%

\algdef{SE}[FUNCTION]{Function}{EndFunction}%
   [2]{\algorithmicfunction\ \textproc{#1}\ifthenelse{\equal{#2}{}}{}{(#2)}}%
   {\algorithmicend\ \algorithmicfunction}%

\DeclareAcronym{NavMoE}{
short=NavMoE,
long=Navigation via Mixture of Experts}

\begin{document}

\title{\AlgName{}: Hybrid Model- and Learning-based Traversability Estimation for Local Navigation via Mixture of Experts}

% You will get a Paper-ID when submitting a pdf file to the conference system
% \author{\\Botao He*, Amir Hossein Shahidzadeh*, Yu Chen*, Jiayi Wu, Tianrui Guan, Guofei Chen, \\ 
% Dinesh Manocha, Glen Chou, Cornelia Fermuller and Yiannis Aloimonos}

\author{
    Botao He $^{\dag \,}$\textsuperscript{1}, 
	Amir Hossein Shahidzadeh $^{\dag}$ \textsuperscript{1}, 
	Yu Chen$^{\dag}$\textsuperscript{2},
    Jiayi Wu\textsuperscript{1},
    Tianrui Guan\textsuperscript{1},
    Guofei Chen\textsuperscript{2}, \\
    Howie Choset\textsuperscript{2},
    Dinesh Manocha\textsuperscript{1},
    Glen Chou\textsuperscript{3},
    Cornelia Fermuller\textsuperscript{1},
	and Yiannis Aloimonos\textsuperscript{1}        
	\thanks{1 Department of Computer Science, University of Maryland, MD 20742.} 
	\thanks{2 Robotics Institute, Carnegie Mellon University, PA 15213-3890.}
    \thanks{3 Georgia Institute of Technology, GA 30332.}
    % \thanks{\textbf{${\dag}$ Equal contribution.}}
	\thanks{${\dag}$ Equal contribution. \ Email: {\tt\small \{botao, jyaloimo\}@umd.edu}}
}

% \author{Anonymous for Review}

%\author{\authorblockN{Michael Shell}
%\authorblockA{School of Electrical and\\Computer Engineering\\
%Georgia Institute of Technology\\
%Atlanta, Georgia 30332--0250\\
%Email: mshell@ece.gatech.edu}
%\and
%\authorblockA{Twentieth Century Fox\\
%\and
%\authorblockN{James Kirk\\ and Montgomery Scott}
%\authorblockA{Starfleet Academy\\
%San Francisco, California 96678-2391\\

% avoiding spaces at the end of the author lines is not a problem with
% conference papers because we don't use \thanks or \IEEEmembership

% \author{\authorblockN{Michael Shell\authorrefmark{1},
%Homer Simpson\authorrefmark{2},
%Montgomery Scott\authorrefmark{3} and
%Eldon Tyrell\authorrefmark{4}}
%Telephone: (800) 555--1212, Fax: (888) 555--1212}
%\authorblockA{\authorrefmark{4}Tyrell Inc., 123 Replicant Street, Los Angeles, California 90210--4321}}

\maketitle

\begin{abstract}
% Robot navigation finds feasible trajectories among traversable areas, which often achieved by analyzing the traversability.
% This paper focuses on traversability analysis for robot navigation, which is generally solved in two ways: analytical approaches based on geometric computation and data-driven approaches based on texture analysis from images. 
% This paper explores sensor data based traversability estimation for robot navigation, typically addressed through two ways: analytical approaches using geometric computation or data-driven methods based on image texture analysis.
This paper explores traversability estimation for robot navigation. 
% Classical approaches typically address this problem based on geometric or physical models.
% These methods offer provable motion safety and efficiency but lack cognitive reasoning capabilities.
% With the rapid development of robot learning, data-driven approaches show strong reasoning ability in traversability estimation based on input image texture analysis.
% However, learning-based traversability estimation still suffer from limited safe guarantee and generalizability.
% Recent works aim to leverage the complementary strengths of both paradigms but face challenges related to efficiency, varying input modalities, and differing pipeline structures.
A key bottleneck in traversability estimation lies in efficiently achieving reliable and robust predictions while accurately encoding both geometric and semantic information across diverse environments.
We introduce Navigation via Mixture of Experts (\AlgName{}), a hierarchical and modular approach for traversability estimation and local navigation.
\AlgName{} combines multiple specialized models for specific terrain types, each of which can be either a classical model-based or a learning-based approach that predicts traversability for specific terrain types.
\AlgName{} dynamically weights the contributions of different models based on the input environment through a gating network.
Overall, our approach offers three advantages:
First, \AlgName{} enables traversability estimation to adaptively leverage specialized approaches for different terrains, which enhances generalization across diverse and unseen environments.
% It's modular architecture makes our approach easily extendable with new experts. 
% It is structured around multiple hybrid modeling modules, each adhering to two consistent design principles:
% First, to enhance the inference efficiency, we introduce the lazy gating mechanism that minimizes the number of experts queried during inference. It is training-free and offers provable optimality in both efficiency and effectiveness.
Second, our approach significantly improves efficiency with negligible cost of solution quality by introducing a training-free lazy gating mechanism, which is designed to minimize the number of activated experts during inference.
% Second, to eliminate the requirement for expensive multi-sensory data and accommodates non-differentiable analytical algorithms for end-to-end training, we propose the two-stage training with pseudo-labeling. It utilize vast human-labeled single-sensor data for pre-training, then generate pseudo labels based on outputs from different experts. In this way, we directly use pseudo labels to supervise the gating policy, thus not requiring experts to be differentiable.
% Second, to avoid costly multi-sensory data and integrate non-differentiable algorithms, we propose a two-stage training approach with pseudo-labeling. We first pretrain the gate on large-scale single-sensor data, then generate pseudo labels from multi-sensory experts to supervise the gating policy. Since we directly supervise the gating policy without passing gradient through experts, we can achieve end-to-end training without requiring differentiable experts.
Third, our approach uses a two-stage training strategy that enables the training for the gating networks within the hybrid MoE method that contains nondifferentiable modules.
% : pretraining the gate on single-sensor data, then using multi-sensory expert pseudo labels to supervise the gating policy for end-to-end training.
% Second, to avoid costly multi-sensory data and integrate non-differentiable algorithms, we propose a two-stage training method with pseudo-labeling. We first pretrain the gate on large-scale single-sensor data. Then, we generate pseudo labels from multi-sensory experts to supervise the gating policy directly, enabling end-to-end training without requiring differentiable experts.
% we propose the two-stage training method, which accommodates non-differentiable analytical algorithms for end-to-end training. 
% The two-stage training method accommodates non-differentiable analytical algorithms for end-to-end training. The modular architecture makes our framework easily extendable with new experts. 
% Extensive experiments on diverse datasets and in real-world scenarios validate that \AlgName{} substantially improves corss-domain generalization while reducing computational cost by \note{XX} without sacrificing path planning quality.
% \yu{Some results to show efficacy}
Extensive experiments show that \AlgName{} delivers a better efficiency
and performance balance than any individual expert or full ensemble across different domains, improving cross-domain generalization and reducing average computational cost by $81.2\%$ via lazy gating, with less than a $2\%$ loss in path quality.

\end{abstract}

\IEEEpeerreviewmaketitle

\section{Introduction}

% \comment{This paragraph discuss what is wrong with learning-based modeling in Robotics}

% \tian{You need to motivate the problem you need to solve}

% \tian{In terms of paragraph, 1) from general local navigation problem -> traversability estimation 2) mention some traditional trav estimation problem -> but their issue is limited semantic information and could not handle complex environment 3) therefore, learning-based methods are introduced, ... -> but they did not scale well and generalizability mind be a issue for untrained / unseen environment. Also a heavy model might be costly (think of your advantage and add those issues in different part of the intro to motivate your method) 4) some of the hybrid method, use weighted sum and is not flexible, not modulated, usually use 1 traditional and 1 learning-based, need to manually tune in different environment.}

We address traversability estimation for safe and efficient robot navigation—a core technique that evaluates how easily a robot can move through an environment, providing key heuristics for path planning~\cite{lidar_nav}.
Classical approaches typically rely on rule-based models based on geometric analysis~\cite{geo1, geo_simple, geo2, geo_trav}.
These approaches extract geometric features such as slope, height variation, and obstacles from sensor inputs and estimate traversability using manually designed heuristics, including terrain roughness thresholds, elevation gradients, and obstacle proximity rules.
% \yu{we need more related works here.}
% By incorporating terrain curvature and robot dynamics into a trajectory optimization framework \cite{xu2023efficient, li2025seb}, Xu et al. enable robots to navigate uneven and complex terrains safely and efficiently.
Such model-based approaches are highly interpretable and allow for the straightforward incorporation of hard constraints on robot motion such as collision avoidance through physical and geometric rules.
As a result, they offer provable safety guarantees and strong generalization across different environments.
Additionally, model-based methods are typically lightweight and computationally efficient \cite{zhang2020falco}. This makes them well-suited for resource-constrained mobile platforms.
However, model-based approaches often struggle in complex or ambiguous environments, where geometric and physical features may be noisy or incomplete~\cite{kpconv, wu2024ptv3}.
They also struggle in understanding high-level features such as semantic information, which limits their capacity to anticipate hidden risks (e.g., recognize and follow traffic signs) or results in over-conservative behaviors.
% However, they are usually not learnable, and struggle to perform high-level reasoning, such as semantic understanding than data-driven approaches.

\begin{figure}[t]
  % \vspace{-0.3cm}
  \begin{center}
    \includegraphics[width=0.45\textwidth]{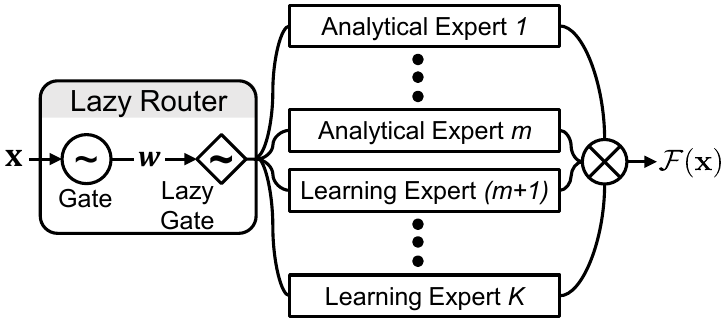}
  \end{center}
  \captionsetup{font={small}}
  \caption{A basic block of the \AlgName{}. It fuses the output of different experts through a learned router.}
    \label{fig:basicblock}
    \vspace{-0.7cm}
\end{figure}

\begin{figure*}[t!]
	% \vspace{0.0cm}
    \centering
	\includegraphics[width=0.95\textwidth]{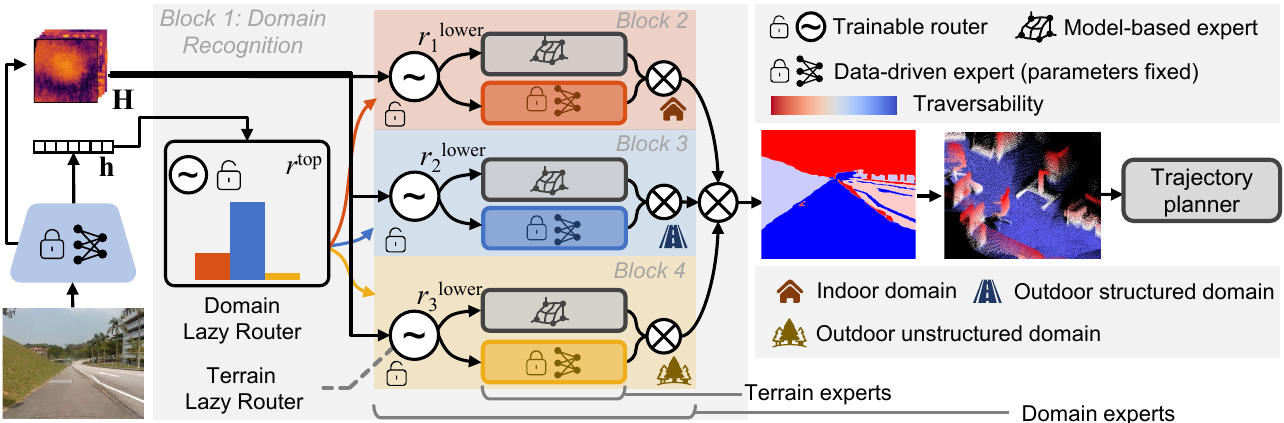}
	% \vspace{-0.6cm}
	\captionsetup{font={small}}
	\caption{
		\textbf{Overview of the hierarchical and modular design.} Given sensor input, the domain router first routes it to the most suitable terrain router. This terrain router then computes a pixel-wise weight map indicating each expert's confidence and queries the terrain experts iteratively. Each expert generates its traversability map, and these outputs are fused according to their weights to form the final estimation. All routers employ the proposed Lazy Gating mechanism. All blocks follow the same modular design illustrated in \cref{fig:basicblock}.}
	\label{fig:method}
	\vspace{-0.6cm}
\end{figure*}

With the rapid development of machine learning, learning-based approaches have become increasingly popular for traversability estimation \cite{tns, vinet, graspe, guan2021ganav, segformer}. 
% \yu{details of these methods here.}
% By classifying terrain types based on semantics understanding \cite{tns, roth2024viplanner} or leveraging self-supervised learning to estimate traversability from multiple sensors \cite{10160856, mattamala2024wild}, these methods demonstrate effective performance in social and off-road navigation.
Compared to model-based approaches, learning-based methods offer significant advantages in unstructured environments.
They also excel at interpreting high-level semantic information in the environment, such as recognizing terrain categories, obstacles, and contextual features that are critical for navigation decisions.
Despite these advantages, learning-based approaches face challenges in providing formal safety guarantees. This is because they often lack interpretability, which makes it difficult to incorporate explicit safety constraints during deployment.
% Moreover, dealing with domain shift remains a bottleneck for learning-based approaches, as models trained in one setting may fail to generalize to others with different terrain types~\cite{hsieh2023distilling, gunasekar2023textbooks, fu2023specializing}.
Also, cross-domain generalization for a single model remains challenging.
% as the recent works \cite{hsieh2023distilling, gunasekar2023textbooks, fu2023specializing} show that advanced foundation models trained on vast cross-domain datasets can be outperformed by smaller specialized models in specific tasks.
% Combining specialized experts with different expertise in a unified structure offers a promising solution for building reliable and efficient robotic systems.
% Their performance is also easily affected by the gap between the training and testing domains, which further limits their generalizability and reliability. 

% \begin{figure}[t!]
% 	% \vspace{0.0cm}
% 	\centering
% 	\includegraphics[width=0.5\linewidth]{Figures/BasicBlock.pdf}
% 	% \vspace{-0.6cm}
% 	% \captionsetup{font={small}}
% 	\caption{
% 		A basic building block of the hybrid modeling.
% 	}
% 	\label{fig:basicblock}
% 	\vspace{-0.5cm}
% \end{figure}
% Combining these two categories in a unified structure offers a promising solution for building reliable and efficient robotic systems.
Combining experts with different expertise is a effective approach in ML. Classical ensemble learning methods like \cite{ganaie2022ensemble} or Mixture of Experts (MoE) \cite{jacobs1991adaptive, wang2023fusing} often leads to performance improvements.
Recent works aim to address the above challenges by integrating model-based and learning-based methods into a unified framework.
\cite{zhu2024learning} encodes images and point clouds into a shared 3D map for geometric analysis and traversability prediction but depends heavily on high-quality multi-sensory data for training.
\cite{haddeler2022traversability} serially connect both paradigms using a human-designed probing strategy, but lacks cross-domain flexibility and is inefficient as both methods must always run. 
\cite{sebastian2019traversability} utilizes physics-informed learning to improve generalization through simulation, but the domain gap remains a challenge. 
% \cite{zhu2024learning} encodes images and point clouds into a shared 3D map for geometric analysis and traversability prediction but depends heavily on high-quality multi-sensory data.
% \cite{haddeler2022traversability} links analytical and learning methods via a human-designed probing strategy, but lacks cross-domain flexibility and efficiency.
% \cite{sebastian2019traversability} uses physics-informed learning to enhance generalization in simulation, yet the domain gap persists.
% While these studies provide valuable insights, they do not fully resolve the key challenges, highlighting the need for a more effective framework that balances safety, efficiency, and generalization.
However, these hybrid approaches still face significant challenges for traversability estimation:
1) It is difficult to determine the contribution of each method, as their performance can vary widely across domains.
2) Hybrid networks often require extensive expensive synchronized multi-sensory data for training.
3) It is often difficult to jointly update the components to enable better cooperation among them. This is because learning-based methods rely on gradient-based optimization, while model-based methods are usually nondifferentiable, as they often involve discrete decision rules or search-based operations that break gradient flow.
% Recent works on differentiable algorithm networks \cite{karkus2019differentiable} aim to encode the parameters of model-based algorithms within a network to enable differentiability. However, these approaches encounter challenges related to scalability and the smoothness of the optimization landscape.

\textbf{Main Results: }
We propose Navigation via Mixture of Experts (\AlgName{}), a hybrid analytical–learning approach designed for traversability estimation.
\AlgName{} unifies both model-based and learning-based approaches by incorporating them as modular components within a hierarchical Mixture of Experts (MoE) architecture.
A basic building block is shown in Fig. \ref{fig:basicblock} and the overall architecture is shown in Fig. \ref{fig:method}.
The novelties of our work include: \textbf{1)} A hierachical MoE approach that performs expert selection at two levels: A high-level router coarsely identifies relevant domains, and a low-level router determines pixel-wise contribution of each expert to the final traversability prediction. \textbf{2)} An expert pruning approach based on the lazy gating mechanism to improve the efficiency of \AlgName{}. The core idea is to only activate an expert when it can lead to significant changes to the robot path solution. 
Instead of calling the best expert, lazy gating calls fast yet effective experts first and adds others only when needed, keeping the accuracy while improving efficiency. 
% It is proven to be able to always find a $\delta$-optimal trajectory for a given set of experts.
We theoretically prove that the path solution obtained using only a subset of experts differs from the solution using all experts by \textit{at most a bounded margin} in terms of solution quality.
\textbf{3)} A two-stage training approach that improves the performance of \AlgName{} with limited high quality multi-sensory data. The first stage involves pre-training on large-scale, weakly-labeled data to learn generalizable representations. It is then followed by a fine-tuning phase using smaller, accurately-labeled datasets to refine task-specific performance.

Extensive experiments in real world and diverse datasets across different domains demonstrate that \AlgName{} achieves $35.3\%$ higher average accuracy in traversability map estimation compared to \textbf{any individual expert}, both purely geometric and learning-based methods.
Moreover, \AlgName{} reduces computational overhead in average by $86.2\%$ compared to our method without lazy gating, with path quality drop within $0.02$. We will open source our code to facilitate future research. We believe the proposed hybrid modeling principle is generic and we expect its application in other robotics tasks.

\section{Hierarchical and Modular Mixture-of-Experts}

An overview of \AlgName{} is shown in Fig.~\ref{fig:method}. Given a RGB image $\mathbf{I} \in \mathbb{R}^{H \times W \times 3}$ and a depth image $\mathbf{D} \in \mathbb{R}^{H \times W}$, \AlgName{} predicts a traversability map $\mathcal{T} \in \mathbb{R}^{H \times W}$ and then computes a feasible robot trajectory $\tau\in \mathbb{R}^{3\times l}$, where $l$ is the trajectory length. 
% Specifically, each pixel in the traversability map $\mathcal{T}_{uv} \in [0, 1]$ indicates the likelihood of being traversable and $\tau = \{\mathbf{x}_1, \mathbf{x}_2, \ldots , \mathbf{x}_T\}$ is a discrete path representing a sequence of feasible robot states. 
\AlgName{} introduces two levels of specialization: \textit{domain experts} $\mathbf{E}_m^{\rm Domain}$ with $m=\{1,2,...,M\}$ at the top level and \textit{terrain experts} $\mathbf{E}_{mn}^{\rm Terrain}$ with $n=\{1,2,...,N\}$ within each domain $m$. Notably, $N$ is the number of terrain experts within each domain; in our setup, $M=3$ and $N=2$, as illustrated in Fig.~\ref{fig:method}. 
% Domain-specific experts ensure that input images are first routed to the most relevant high-level domain category, which allows for finer optimization within each domain and reduces disturbance from unrelated domains. Within each domain, terrain-specific expertise further refines predictions by focusing on distinct characteristics of terrain types.
Domain experts route input images to relevant high-level domains for targeted optimization, while terrain experts further refine predictions based on distinct terrain features.

% \vspace{-0.2cm}
\paragraph{Top Level Router: Domain Specialization}
% \vspace{-0.2cm}
% At the top level, \AlgName{} performs a coarse specialization by partitioning the input domain. In our framework, the domains are predefined primarily based on the types of scenarios that the robot may encounter. 
% Specifically, we consider three types of domains: 1) the indoor domain consists of indoor scenarios such as hallways, rooms, or warehouses, 2) the structured outdoor domain consists of road scenarios with structured features such as lanes, sidewalks, or crosswalks, and 3) the unstructured outdoor domain consists of complex and off-road scenarios such as forests, rocky terrain, or grassy fields.
At the top level, \AlgName{} performs coarse specialization by partitioning the input domain based on scenario types. Within the scope of this paper, we define three domains: 1) indoor, including hallways, rooms, and warehouses; 2) structured outdoor, featuring roads with lanes, sidewalks, and crosswalks; and 3) unstructured outdoor, encompassing off-road settings like forests, rocky terrain, and grassy fields.
% This mechanism is based on common sense \yu{Is this a common sense? I might need citations here} that different types of scenario cause different environment characteristics and distributions of input features, which will significantly affect navigation strategies. 

We formulate domain partitioning as an image classification problem.
\AlgName{} first extracts a dense feature map $\mathbf{H} \in \mathbb{R}^{H\times W \times d_1}$ and a feature vector $\mathbf{h} \in \mathbb{R}^{d_2}$ from the input images using a pre-trained encoder $\mathcal{E}$: 
\vspace{-2mm}
\begin{equation}
    \mathbf{H}, \mathbf{h} = \mathcal{E}(\mathbf{I})
\end{equation}
where we use DINOv2 \cite{oquab2023dinov2} as the encoder $\mathcal{E}$.
% \footnote{$d_1\neq d_2$ in general} 
After that, the top-level router $r^{\rm top}: \mathbb{R}^{d_2} \rightarrow \mathbb{R}^M$ takes $\mathbf{h}$ as input for domain classification. The output of the top-level router is a normalized vector $\mathbf{w}_m \in \mathbb{R}^{M\times1}$, which indicates the predicted possibilities for each type of domain: $\mathbf{w}_m = r^{\rm top} (\mathbf{h})$.
Then, based on the predicted weights $\mathbf{w}_m$, the domain experts $\mathbf{E}^{\rm Domain}$ are queried using the proposed lazy gating algorithm (detailed in \cref{sec:router}).
% Then, $\mathbf{w}^{\rm D}$ is utilized to query the domain-specific experts $\mathbf{E}_k^{\rm D}$. 
% \begin{equation}
%     \mathbf{w}^{\rm Domain} = r^{\rm top} (\mathbf{h})
% \end{equation}
% where $\mathbf{w}^{\rm Domain}$ serve as weights for domain-specific experts $\mathbf{E}_k^{\rm Domain}$. 

% \vspace{-0.2cm}
\paragraph{Lower Level Router: Terrain Specialization}
Within the selected domain, we implement a lower-level expert routing mechanism. Specifically, a router $r^{\rm lower}$ assigns terrain experts $\mathbf{E}^{\rm Terrain}_{mn}$ to each pixel based on a pixel-wise terrain partition of the input image.
    Unlike the top-level coarse classification that assigns the whole input to several domain categories, $r^{\rm lower}: \mathbb{R}^{H\times W\times d_1} \rightarrow \mathbb{R}^{H\times W\times N}$ performs fine-grained segmentation by predicting pixel-wise terrain expert preferences. It takes feature map $\mathbf{H}$ as input and predicts a weight map of expert weights $\mathbf{W} \in \mathbb{R}^{H \times W \times N}$:
\vspace{-1mm}
\begin{equation}
    \mathbf{W} = r^{\rm lower} (\mathbf{H})
    \label{eq:weight_map}
\end{equation}
Therefore, weight map for each expert can be expressed as $\mathbf{W}_n \in \mathbb{R}^{H \times W}$.
% \vspace{-1mm}
Each terrain expert $\mathbf{E}^{\rm Terrain}_{mn}$ is implemented as either a neural network or a model-based algorithm (details in \cref{sec:exp}). These experts take RGB or depth images as input and produce pixel-aligned traversability maps $\mathcal{T}_{mn} \in \mathbb{R}^{H\times W}$.
Formally, an expert can be expressed as a mapping $\mathbf{E}_{mn}^{\rm Terrain}:\mathbb{R}^{H\times W\times C} \rightarrow \mathbb{R}^{H\times W}$, where $C=3$ if the input is an RGB image and $C=1$ if the input is a depth image. The final $\mathcal{T}$ is computed as a two-level weighted combination of expert predictions:
\vspace{-1mm}
{%\small
\begin{equation}
    \mathcal{T} = \sum_{m=1}^M \left( \mathbf{w}_m \cdot \sum_{n=1}^N \left( \mathbf{W}_n^{\rm lower} \odot \mathcal{T}_{mn} \right) \right),
\end{equation}
}

\noindent where $\odot$ refers to elementwise multiplication. The resulting $\mathcal{T}$ is then passed to a motion planner to compute an optimal trajectory $\tau^*$.
Here we adopt the primitive-based planner in \cite{zhang2020falco}.
Defining the candidate motion primitives set for the traversability map $\mathcal{T}$ as $S(\mathcal{T}) = \{\tau_1,\tau_2,...,\tau_l\}$,
we formulate the trajectory optimization problem as:
\vspace{-2mm}
% \begin{equation}
%     \tau^* = \argmin_{\tau \in S(\mathcal{T})} \sum_{i=1}^{l} 
%      J_\textrm{trav}\left(\tau_i, \mathcal{T}\right) + \lambda  J_\textrm{dis} \left( \tau_i, p_\textrm{goal} \right).
%     \label{eq:planner}
% \end{equation}
\begin{equation}
f(\mathcal{T}, p_{\mathrm{goal}})
\triangleq
\operatorname*{arg\,min}_{\tau \in S(\mathcal{T})}
\sum_{i=1}^{\ell}
\Big( J_{\mathrm{trav}}(\tau_i,\mathcal{T})
+ \lambda\, J_{\mathrm{dis}}(\tau_i, p_{\mathrm{goal}}) \Big).
    \label{eq:planner}
\end{equation}
$J_\textrm{trav}\left(\tau_i, \mathcal{T}\right)$ represents the traversability cost of the path $\tau_i$ on $\mathcal{T}$, $J_\textrm{dis} \left( \tau_i, p_\textrm{goal} \right)$ represents the cost of $\tau_i$ with respect to reaching the goal $p_\textrm{goal}$. Finally, $\tau^* = f(\mathcal{T}, p_{\mathrm{goal}})$.

Specifically, let $J_{\text{trav}}(\tau, \mathcal{T})$ denote the traversability cost of a 3D trajectory $\tau$ with respect to a 2D traversability map $\mathcal{T}$. Utilizing the depth information, we first project $\tau = \{q_1, q_2, \dots, q_l\}$, where $q_i \in \mathbb{R}^3$, onto the 2D plane of $\mathcal{T}$, yielding $\hat{\tau} = \{p_1, p_2, \dots, p_l\}$, where each $p_i = (x_i, y_i) \in \mathbb{R}^2$ corresponds to a waypoint on the map.
The traversability cost is then computed by summing the map values at each projected waypoint: $ J_{\text{trav}}(\tau, \mathcal{T}) = \left. \sum_{p_i \in \hat{\tau}} \left(1 - \mathcal{T}\left(p_i\right)\right) \right/ |\tau|.$
% \begin{equation}
%  J_{\text{trav}}(\tau, \mathcal{T}) = \left. \sum_{p_i \in \hat{\tau}} \left(1 - \mathcal{T}\left(p_i\right)\right) \right/ |\tau|.
% \end{equation}
Notably, since $J_{\text{trav}}(\tau, \mathcal{T})$ is defined as a cost, lower values indicate better traversability.

For $J_\textrm{dis} \left( \tau, p_\textrm{goal} \right)$, let $\tau = \{q_1, q_2, \dots, q_l\}$ denote a candidate 3D trajectory and $p_{\text{goal}} \in \mathbb{R}^3$ be the goal position. Let $p \in \mathbb{R}^3$ be the current robot position. Define the distance-based cost as: $J_{\text{dis}}(\tau, p_{\text{goal}}) = \left. 1-
 (\|p - p_{\text{goal}}\| - \min_{q_j \in \tau} \|q_j - p_{\text{goal}}\|) \right/ H ,$
% \begin{equation}
% J_{\text{dis}}(\tau, p_{\text{goal}}) = 1-
% \frac{ \|p - p_{\text{goal}}\| - \min_{q_j \in \tau} \|q_j - p_{\text{goal}}\| }
%      { H },
% \end{equation}
where $H$ denotes the maximum planning distance, which means that for all $q_j$ in $\tau$, $\|q_j - p\| \leq H$. $H$ is set as $2m$ for indoor, $4 m$ for unstructured outdoor and $8m$ for unstructured outdoor environment. The result lies in the range of $[0,1]$.

% The formulation of $J_{trav}\left(\tau_i, \mathcal{T}\right)$, $J_{dis} \left( \tau_i, p_{goal} \right)$, and $S(\mathcal{T})$ is detailed in \yu{XXX}.
% where $R_{trav}(\tau_i, \mathcal{T})$ gives the traversability reward for path $\tau_i$ on $\mathcal{T}$, and $R(\tau_i, p_{goal})$ denotes the reward of the path $\tau_i$ to the goal $p_{goal}$. The detailed definition of $R_{trav}(\tau_i, \mathcal{T})$ and $R(\tau_i, p_{goal})$ is detailed in \note{Appendix. XXX}

% \begin{equation}
%     \tau = {\rm Planner}(\mathcal{T})
%     \label{eq:planner}
% \end{equation}

% In the following sections, we use $C(\mathcal{T})$to denote the cost of the planned trajectory $\tau$ based on the given traversability map.

% \yu{We might need to formulate the planner, tell people the constraints and objective function, we need this for the proof part}

\section{Lazy Gating Mechanism for Expert Pruning}\label{sec:router}
\subsection{Lazy Gating Mechanism}
\begin{figure*}[t!]
	% \vspace{0.0cm}
	\centering
	\includegraphics[width=0.95\textwidth]{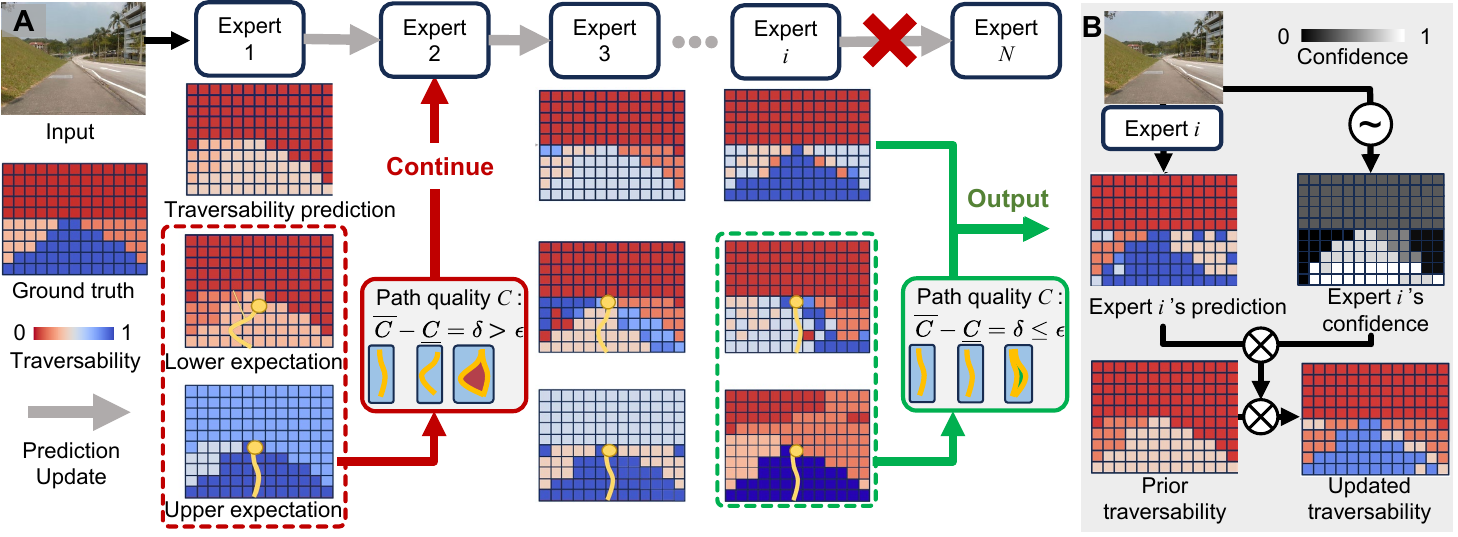}
	% \vspace{-0.6cm}
	\captionsetup{font={small}}
	\caption{
		\textbf{Lazy Gating Mechanism.} \textbf{(A)} Illustration of expert pruning during inference. \textbf{(B)} One iteration of the traversability map update.}
	\label{fig:lazy_gaiting}
	\vspace{-0.5cm}
\end{figure*}

As the number of experts increases, activating all experts can cause severe computational overhead.
To address this challenge, we introduce the lazy gating mechanism as detailed in \cref{alg:lazy_gating}. The key idea is to selectively activate a subset of experts that could significantly impact the path solution. We implement the lazy gating mechanism to both lower and higher level routers.

{\footnotesize
\begin{algorithm}[htbp]
    \small
\caption{Lazy Gating}\label{alg:lazy_gating}
% \vspace{-5mm}
% \begin{multicols}{2}
\begin{algorithmic}[1]
\Function{LazyGating}{}
    \State{Compute $\mathbf{W} \in \mathbb{R}^{H\times W \times N}$ using \cref{eq:weight_map}}
    \State{Construct $\mathcal{Q}_E$ using \cref{eq:expert_queue}}
    \State{$\mathcal{T}\gets$ \texttt{None}, $\mathbf{W}^{\rm prior} \gets 0$}
    \For{$k = 1, 2, \cdots, \mathcal{K}$}
    \State{$E \gets$ \texttt{Dequeue}($\mathcal{Q}_E$) }
    \State{$\hat{\mathcal{T}} \gets E(\mathbf{I}, \mathbf{D})$}
    \If{$k==1$}
    \State{$\mathcal{T} \gets \hat{\mathcal{T}}$, $\mathbf{W}^{\rm prior} \gets \mathbf{W}[n]$}
    \Else
    \State{$\mathcal{T} \gets \frac{\mathcal{T} \cdot \mathbf{W}^{\rm prior} + \hat{\mathcal{T}} \cdot \mathbf{W}[k]}{\mathbf{W}^{\rm prior}  + \mathbf{W}[k]}$}\Comment{\cref{eq:bayesian_update}}
    \State{$\mathbf{W}^{\rm prior} \gets \mathbf{W}^{\rm prior} + \mathbf{W}[k]$}
    \EndIf
    \State{\textit{Done} $\gets$ \texttt{CheckTerm}}
    \If{\textit{Done}}
    \State{\textbf{break}}
    \EndIf
    \EndFor
    \State{\textbf{return} $\mathcal{T}$}
\EndFunction
\vspace{0.5em}
\Function{CheckTerm}{}
\State{Compute $\underline{\mathcal{T}}$ and $\overline{\mathcal{T}}$ using \cref{eq:travers_lower_expec} and \cref{eq:travers_upper_expec}, respectively}
\State{$\overline{C} \gets$ Planner($\underline{\mathcal{T}}$), $\underline{C} \gets$ Planner($\overline{\mathcal{T}}$)}
\If{$\overline{C} - \underline{C} \leq \epsilon$}
\State{\textbf{return} True}
\Else
\State{\textbf{return} False}
\EndIf
\EndFunction    
\end{algorithmic}
% \end{multicols}
% \vspace{-3.5mm}
\end{algorithm}}

For the case of the lower router, we define the cost function $\Phi(\mathbf{E}_{mi}^\textrm{Terrain}) = (1 - ||\mathbf{W}_i||_1 / \sum_{n=1}^{N} ||\mathbf{W}_n||_1) \cdot \phi(\mathbf{E}_{mi}^\textrm{Terrain})$, where $||\cdot||_1$ is the pixel-wise sum of the weight map and $\phi(\cdot)$ indicates the computational cost of an expert, here measured with FLOPS (Floating Point Operations). We consider the lower router case here; however, the same logic can be applied to the top-level router as well. 
We begin by constructing an expert queue based on $\Phi(\mathbf{E}_{mi}^\textrm{Terrain})$:
% \begin{equation}
% \begin{split}
% \mathcal{Q}_E =& \left\{ (E_1, P_{E_1}), \cdots , (E_k, P_{E_k}), \cdots, (E_K, P_{E_K}) \right\} \\
%     s.t. \quad & \mathcal{S}(E_1) \leq \mathcal{S}(E_2) \leq \cdots \leq \mathcal{S}(E_K)
% \end{split}
% \label{eq:expert_queue}
% \end{equation}
% \vspace{-2mm}
\begin{equation}
\label{eq:expert_queue}
\begin{aligned}
\mathcal{Q}_E &= \left\{ E_1, \dots, E_k, \dots, E_{\mathcal{K}} \right\} \\
\text{s.t.}\quad & \Phi(E_1) \le \Phi(E_2) \le \dots \le \Phi(E_{\mathcal{K}}).
\end{aligned}
\end{equation}
Here, $\mathcal{K} = K$ defined in Fig. \ref{fig:basicblock}; however, we adopt a different notation to represent the reordering process used to construct the queue.
Intuitively, if experts have similar weights, we prioritize those with higher computational efficiency; if one has a dominant weight, we prioritize even if it has lower efficiency.
As shown in \cref{fig:lazy_gaiting} \textcolor{anotBlue}{A}, the lazy gating algorithm processes experts sequentially, iterating from the front of the queue to the rear. During each iteration, the next expert is dequeued and queried to provide a traversability prediction. The internal processing of each expert is shown in \cref{fig:lazy_gaiting} \textcolor{anotBlue}{B}: At iteration $k$, the algorithm maintains a prior traversability map $\mathcal{T}_{k-1}$, which aggregates the predictions from the previous $k-1$ experts. The $k$th expert predicts a traversability map $\hat{\mathcal{T}}_k$ while the gating network estimates the current expert’s confidence as weight $\mathbf{W}_k$. The current expert’s output $\hat{\mathcal{T}}_k$ is then fused into the overall map:
{%\small
\begin{equation}
    \mathcal{T}_{k} = \left.\left(\mathcal{T}_{k-1} \odot \textstyle\sum_{j=1}^{k-1}\mathbf{W}_j + \mathcal{T}_k \odot \mathbf{W}_k \right) \right/ \textstyle\sum_{j=1}^{k}\mathbf{W}_j,     \label{eq:bayesian_update}
\end{equation}
}
\vspace{-0.3cm}

\noindent \looseness-1where the division is performed elementwise. The updated traversability map $\mathcal{T}_{k}$ is then used as the new prior for the next iteration. The planner will take $\mathcal{T}_{k}$ as input to compute the path $\tau_k$ and path cost $C(\tau_k)$ using \cref{eq:planner}. Given $C(\tau_k)$ as prior, the expected cost of the final path $C(\tau_K)$ will be constrained to lie within an interval after any admissible update from expert $k+1$ to expert $K$; we discuss how to get the bounds of this interval in the following proposition. To reduce the computation overhead, we prune the rest of the experts if they do not significantly change the traversability output by the threshold $\epsilon$, as detailed in lines 18-25 of Alg. \ref{alg:lazy_gating}.

\begin{proposition}[Convergence Bound on Path Cost Under Incremental Traversability Updates]
    Let $\mathcal{T}_k$ be the traversability map after $k$ aggregative updates following \cref{eq:bayesian_update}, and let $C(\mathcal{T}_k) \doteq C(\tau^*)$ denote the path cost computed using $\mathcal{T}_k$. 
% $C(\mathcal{T}_k)$
Define upper and lower expectation maps based on the remaining uncertainty in traversability ($[0, 1]$) from step $k + 1$ to $K$ as:
{%\small
\begin{subequations}
    \begin{equation}
    \begin{aligned}
        \underline{\mathcal{T}_k} 
        &= \left. \left( \mathcal{T}_k \odot \textstyle\sum_{j=1}^k \mathbf{W}_j 
            + 0 \cdot \textstyle\sum_{j=k+1}^K \mathbf{W}_j \right) \right/ \textstyle\sum_{j=1}^K \mathbf{W}_j \\
        &= \mathcal{T}_k \odot \textstyle\sum_{j=1}^k \mathbf{W}_j 
           \left/ \textstyle\sum_{j=1}^K \mathbf{W}_j \right. ,
    \end{aligned}
    \label{eq:travers_lower_expec}
    \end{equation}
    
    \vspace{-13pt}
    \begin{equation}
        \overline{\mathcal{T}_k} = \left. \left( \mathcal{T}_k \odot \textstyle\sum_{j=1}^k \mathbf{W}_j + 1 \cdot \textstyle\sum_{j=k+1}^K \mathbf{W}_j \right) \right/ \textstyle\sum_{j=1}^K \mathbf{W}_j ,
        \label{eq:travers_upper_expec}
    \end{equation}
\end{subequations}
}

\noindent where $\mathbf{W}_j$ denotes the weight related to the $j$-th update predicted by the routers, and $K$ is the total number of updates.
Intuitively, the lower expectation map assumes that all remaining (unexecuted) experts label the environment as untraversable, while the upper expectation map assumes they label it as fully traversable.
Then, the deviation of the current path cost from the final one is bounded as
\begin{equation}
\small
    \left| C(\mathcal{T}_k) - C(\mathcal{T}_K) \right| \leq \delta_k,
    \quad \text{where} \quad
    \delta_k := C\left(\underline{\mathcal{T}_k}\right) - C\left(\overline{\mathcal{T}_k}\right),
    \label{eq:delta-optimality}
\end{equation}
and the bound $\delta_k$ is non-increasing with $k$, i.e., 
\begin{equation}
    \delta_k \geq \delta_{k+1} \quad \forall k \in \{1, \ldots, K-1\}.
    \label{eq:delta_monotonic_decrease}
\end{equation}
\end{proposition}

The full proof is provided in \cref{sec:appendix-proof}.

\begin{figure}[t]
	% \vspace{0.0cm}
	\centering
	\includegraphics[width=0.45\textwidth]{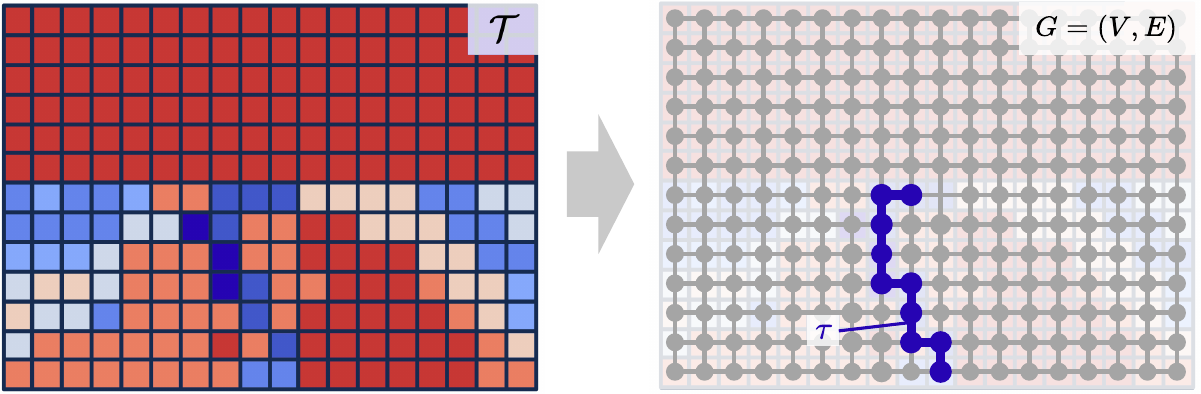}
	% \vspace{-0.6cm}
	\captionsetup{font={small}}
	\caption{$\mathcal{T}$ is modeled as an undirected graph $G = (V, E)$.  A path is modeled as a sequence of vertices $\tau$ connected via $G$.}
	\label{fig:map2graph}
	\vspace{-0.8cm}
\end{figure}

The proposition implies that the path cost converges within a provably shrinking bound as more expert updates are aggregated.
It guarantees that the path cost computed from a partially updated traversability map, obtained from only a subset of experts, deviates from the final cost computed using all experts by a bounded, computable margin.

The proposition establishes that activating all experts is unnecessary to obtain a satisfactory path solution, which provides theoretical justification for our lazy gating approach. Specifically, if the convergence bound $\delta$ falls below a predefined threshold $\epsilon$, the path cost is considered sufficiently stable, and the influence of remaining expert predictions is deemed negligible. This enables early termination of the update loop. This improves computational efficiency without significantly compromising path quality, as the traversability map has already converged within an acceptable tolerance.

% Guofei: Intuitively, the lower expectation map assumes every unexecuted expert marks the observation untraversable, whereas the higher expectation map assumes they all mark it traversable. This proposition states that the cost of the calculated path from the traversability estimate from a subset of experts is tightly bounded. This bound can be calculated by querying the cheap path planner with the lower and upper expectations map of the current partial traversability estimate.

% \vspace{-0.7cm}
\subsection{Proof of Convergence Bound} \label{sec:appendix-proof}

To support the proof, we model a traversability map $\mathcal{T}$ as an undirected graph $G = (V, E)$, where each vertex $v \in V$ corresponds to a pixel in $\mathcal{T}$ that is assigned a scalar traversability value.
% (as shown in \cref{fig:map2graph}). 
An edge $(v_i, v_j) \in E$ exists if the pixels are adjacent, capturing local connectivity. A path through the environment is defined as a sequence of vertices $\tau = [v_1, v_2, \cdots , v_t, \cdots, v_T]$ connected via $G$.
The path planner seeks an optimal path $\tau^*$ by minimizing an objective function subject to feasibility constraints. 
We formulate the path planning problem within graph $G$ as a constrained optimization problem $\mathcal{P}$:
\begin{equation}
\small
    \begin{aligned}
        \mathcal{P}(\mathcal{T}): & \quad C(\mathcal{T}) = \min \{f(\tau; \mathcal{T}) : \tau \in S(\mathcal{T})\} \\
        &\quad{\rm s.t.} \quad  f(\tau; \mathcal{T}) = \sum_{t=1}^T (1 - \mathcal{T}(v_t))^2 + f'(\tau) 
        \\ & \quad \quad \quad
        S(\mathcal{T}) = \{ \tau | \mathcal{T}(v_t) \geq \theta \ \forall t, v_1 = v_{\rm src} \}
    \end{aligned}
    \label{eq:path-problem-formulation}
\end{equation}
where $f(\tau; \mathcal{T})$ is the cost of $\tau$ under traversability map $\mathcal{T}$, and $S(\mathcal{T})$ is the set of feasible paths.
The first term in $f$ penalizes low-traversability paths, and $f'$ indicates auxiliary objectives such as path length, smoothness, or dynamics constraints. 
The definition of the feasible path set $S(\mathcal{T})$ indicates that the traversability of every vertex in the path should be at least of threshold $\theta$.

% The path planner seeks an optimal path $\tau^*$ by minimizing an objective function subject to feasibility constraints. We formulate this as a constrained optimization problem:
% \begin{equation}
%     \begin{aligned}
%         \mathcal{P}(\mathcal{T}): C(\mathcal{T}) &= \min \{f(\tau; \mathcal{T}) : \tau \in S(\mathcal{T})\} \\
%         {\rm s.t.} \quad & f(\tau; \mathcal{T}) = \sum_{t=1}^T (1 - \mathcal{T}(v_t))^2 + f'(\tau) \\
%         &
%         S(\mathcal{T}) = \{ \tau | \mathcal{T}(v_t) \geq \theta \ \forall t, v_1 = v_{\rm src} \}
%     \end{aligned}
%     \label{eq:path-problem-formulation}
% \end{equation}
% where $f(\tau; \mathcal{T})$ is the cost of $\tau$ under traversability map $\mathcal{T}$, and $S(\mathcal{T})$ is the set of feasible paths.
% The first term in $f$ penalizes low-traversability paths, and $f'$ indicates auxiliary objectives such as path length, smoothness, or dynamics constraints. 
% The definition of the feasible path set $S(\mathcal{T})$ indicates that the traversability of every vertex in the path should be at least of threshold $\theta$.

We proceed to prove the proposition in two parts. First, we establish that the bound $\delta_k$ is well-defined. Then, we show that $\delta_k$ monotonically decreases with increasing $k$.

\subsubsection{Validity and Bound Definition.}\label{sec:proof-validity-and-bound}
% By \cref{eq:travers_lower_expec} and \cref{eq:travers_upper_expec}, it is obvious that given any path $\tau$, for any waypoint in the path $v_t$, 
From the definitions of \( \underline{\mathcal{T}_k} \) and \( \overline{\mathcal{T}_k} \) in \cref{eq:travers_lower_expec,eq:travers_upper_expec}, it follows that for any vertex \( v_t \) along a path \( \tau \), we have:
\begin{equation}
    \underline{\mathcal{T}_k}(v_t) \leq \mathcal{T}_k(v_t) \leq \overline{\mathcal{T}_k}(v_t)
    \label{eq:travers_compare}
\end{equation}

The elementwise inequality in \cref{eq:travers_compare} implies that, for any fixed path \( \tau \), the traversability component of the cost function in \cref{eq:path-problem-formulation} is bounded:
\begin{equation}
    f(\tau; \overline{\mathcal{T}_k}) \leq f(\tau; \mathcal{T}_k) \leq f(\tau; \underline{\mathcal{T}_k})
    \label{eq:path-problem-formulation_compare}
\end{equation}

% Given the inequality in \cref{eq:travers_compare} and \cref{eq:path-problem-formulation}, it follows that:
% \begin{equation}
%     S(\underline{\mathcal{T}_k}) \subseteq S(\mathcal{T}_k) \subseteq S(\overline{\mathcal{T}_k})
%     \label{eq:subset_relationship}
% \end{equation}

% The nesting of feasible sets in \cref{eq:subset_relationship}, combined with the elementwise cost bounds in \cref{eq:path-problem-formulation_compare}, implies that $\mathcal{P}(\mathcal{T}_k)$ is a relaxation of $\mathcal{P}(\overline{\mathcal{T}_k})$, and $\mathcal{P}(\underline{\mathcal{T}_k})$ is a further relaxation of $\mathcal{P}(\mathcal{T}_k)$. 
Given the inequality in \cref{eq:travers_compare} and \cref{eq:path-problem-formulation}, it follows that 
$S(\underline{\mathcal{T}_k}) \subseteq S(\mathcal{T}_k) \subseteq S(\overline{\mathcal{T}_k})$. 
The nesting of these feasible sets, combined with the elementwise cost bounds in \cref{eq:path-problem-formulation_compare}, implies that $\mathcal{P}(\mathcal{T}_k)$ is a relaxation of $\mathcal{P}(\overline{\mathcal{T}_k})$, and $\mathcal{P}(\underline{\mathcal{T}_k})$ is a further relaxation of $\mathcal{P}(\mathcal{T}_k)$.

From the definition of problem relaxation, we obtain a corresponding ordering of the optimal path costs:
\begin{equation}
    C(\overline{\mathcal{T}_k}) \leq C(\mathcal{T}_k) \leq C(\underline{\mathcal{T}_k})
    \label{eq:optimal_cost_compare}
\end{equation}

To extend this to the true final $\mathcal{T}_K$, we observe that the elementwise value of $\mathcal{T}_K$ is bounded between $\underline{\mathcal{T}_k}$ and $\overline{\mathcal{T}_k}$:
\begin{equation}
\small
    \underline{\mathcal{T}_k} - \mathcal{T}_K = 
    - \sum_{j=k+1}^K \mathcal{T}_j \cdot \mathbf{W}_j \left/ \sum_{j=1}^K \mathbf{W}_j \right. \leq 0 
     \Rightarrow \underline{\mathcal{T}_k} \leq \mathcal{T}_K
\end{equation}
\begin{equation}
\small
    \mathcal{T}_K - \overline{\mathcal{T}_k} = \left.- \sum_{j=k+1}^K (1-\mathcal{T}_j) \cdot \mathbf{W}_j \right/ \sum_{j=1}^K \mathbf{W}_j \leq 0
    \Rightarrow \mathcal{T}_K \leq \overline{\mathcal{T}_k}
\end{equation}

Therefore, applying the same logic as above, we obtain:
\begin{equation}
    C(\overline{\mathcal{T}_k}) \leq C(\mathcal{T}_K) \leq C(\underline{\mathcal{T}_k})
    \label{eq:optimal_cost_compare_of_K}
\end{equation}

Combining \cref{eq:optimal_cost_compare} and \cref{eq:optimal_cost_compare_of_K}, we conclude:
\begin{equation}
    |C(\mathcal{T}_k) - \mathcal{T}_K| \leq C(\underline{\mathcal{T}_k}) - C(\overline{\mathcal{T}_k})
\end{equation}

This establishes that \( \delta_k = C(\underline{\mathcal{T}_k}) - C(\overline{\mathcal{T}_k}) \) is a valid upper bound on the absolute error between the current and final cost estimates as stated in \cref{eq:delta-optimality}.

\subsubsection{Monotonicity of $\delta_k$}
We now show that the convergence bound \( \delta_k = C(\underline{\mathcal{T}_k}) - C(\overline{\mathcal{T}_k}) \) is non-increasing over iterations.

First, we examine the difference between $\overline{\mathcal{T}_{k+1}}$ and $\overline{\mathcal{T}_k}$:
\begin{equation}
\small
    \overline{\mathcal{T}_{k+1}} - \overline{\mathcal{T}_k} = \left( \hat{\mathcal{T}}_{k+1} - 1 \right) \cdot \frac{\mathbf{W}_{k+1}}{\sum_{j=1}^K \mathbf{W}_j} \leq 0 \Rightarrow \overline{\mathcal{T}_{k+1}} \leq \overline{\mathcal{T}_k}
\end{equation}

Similarly, for the pessimistic estimate we have $\underline{\mathcal{T}_{k+1}} \geq \underline{\mathcal{T}_k}$.
% \begin{equation}
% \small
%     \underline{\mathcal{T}_k} - \underline{\mathcal{T}_{k+1}} = - \hat{\mathcal{T}}_{k+1} \cdot \frac{\mathbf{W}_{k+1}}{\sum_{j=1}^K \mathbf{W}_j} \leq 0 \Rightarrow \underline{\mathcal{T}_{k+1}} \geq \underline{\mathcal{T}_k}
% \end{equation}

Therefore, applying the same logic as in \cref{sec:proof-validity-and-bound}, we obtain:
% \begin{equation}
%     \left. 
%     \begin{array}{r}
%         C(\overline{\mathcal{T}_{k+1}}) \geq C(\overline{\mathcal{T}_k}) \\
%         C(\underline{\mathcal{T}_{k+1}}) \leq C(\underline{\mathcal{T}_k}) 
%     \end{array}
%     \right\}
%     \Rightarrow
%     C(\underline{\mathcal{T}_{k+1}}) - C(\overline{\mathcal{T}_{k+1}}) \leq C(\underline{\mathcal{T}_k}) - C(\overline{\mathcal{T}_k})
% \end{equation}
\begin{equation}
    C(\overline{\mathcal{T}_k}) \leq C(\overline{\mathcal{T}_{k+1}}) \leq
    C(\underline{\mathcal{T}_{k+1}}) \leq C(\underline{\mathcal{T}_k})
\end{equation}

Therefore, the convergence bound satisfies:
\begin{equation}
    \delta_{k+1} = C(\underline{\mathcal{T}_{k+1}}) - C(\overline{\mathcal{T}_{k+1}})
    \leq C(\underline{\mathcal{T}_k}) - C(\overline{\mathcal{T}_k}) = \delta_k
\end{equation}

This establishes that \( \delta_k \) is monotonically non-increasing with \( k \).

\section{Training Strategy}\label{sec: training strategy}
We first pretrain the gating networks on large-scale partially-aligned semantic segmentation datasets using human-prior-based label mappings to encourage generalization. Then, we fine-tune the router and experts end-to-end on small-scale multi-modal data with weak supervision derived from expert-label consistency.
% \note{our experts: indoor: maskformer finetuned on ade20k dataset (one of the most popular segmentation models for realtime indoor semantics segmentation); outdoor-structured: maskformer finetuned on mapiliary vistas dataset (the only model on sota benchmark that has less than 100M params); outdoor: GaNav (current sota on both datasets we are using)}

% We adopt a two-stage training strategy that is commonly applied in previous works on robot learning \yu{add 2–3 citations}. In the first stage, \AlgName{} is pre-trained on multiple datasets whose domains and data structures are only partially aligned with our task. In the second stage, the model is fine-tuned using a smaller, high-quality dataset collected specifically for our robot traversability analysis.
% Furthermore, instead of training the entire pipeline, we only train each gaiting module.

\subsection{Pre-Training on Large-Scale Partially-Aligned Data}

\begin{table}[b]
\centering
\small
\vspace{-0.5cm}
\begin{tabularx}{\linewidth}{c|X|c}
\toprule
\multicolumn{1}{>{\centering\hspace{0pt}}m{0.5\linewidth}|}{\textbf{Classes}} & \textbf{Trav. Cost} & \textbf{Ass. Exp.}  \\ 
\hline\hline
sidewalk, floor, crosswalk, bike lane, etc   & $c_{free}$      & M     \\ 
\hline
gravel, sand, snow, low grass, etc                                       & $c_{mid1}$      & NN                   \\ 
\hline
road, parking area                                                             & $c_{mid2}$      & NN                   \\ 
\hline
terrain (tall grass, bush, dirt)                                                     & $c_{mid3}$      & NN                   \\ 
\hline
person, curb, wall, other obstacles                                                              & $c_{obs}$       & M                   \\
\bottomrule
\end{tabularx}
\captionsetup{font={small}}
\caption{Semantics to assigned expert. M: model-based method; NN: Neural Networks. The definition of traversability cost $c$ follows the prior work \cite{roth2024viplanner}.}
\vspace{-0.2cm}
\label{tab:sem2trav}
\end{table}

The gating networks are first pretrained on large-scale data that contains various datasets with a wide domain distribution to learn the broad and general features. 
% To sufficiently extend the scale of the pre-training data, we introduce various semantic segmentation datasets for related tasks such as autonomous driving, \yu{XXX}, etc. 
These data include 32,000 images from three single-sensory datasets \cite{ade20k, neuhold2017mapillary, RUGD2019IROS}, with each dataset corresponding to a distinct domain.
However, it is impractical to directly use these datasets for training, as they contain only 2D images without synchronized depth information. This limitation makes it difficult to implement certain experts, such as model-based approaches that rely on geometric data.
We solve this problem by mapping semantic or instance segmentation labels to expert selection labels based on human knowledge.
The label mapping encourages model-based approaches to handle terrains with clear geometric features, while assigning terrains where semantic information is more critical to traversability analysis to various neural networks.
An example of the label mapping is shown in \cref{tab:sem2trav}.
We encourage the gating networks to assign higher confidence to model-based experts with terrains such as roads or sidewalks. Meanwhile, higher confidence is encouraged to be assigned to experts based on neural networks on terrains such as grass or water surfaces.

% \begin{equation}
%     equation \ of \ loss
% \end{equation}
Although human priors do not perfectly capture the performance of each expert across domains or terrains, our experiments demonstrate that this pre-training strategy effectively paves the way for the subsequent fine-tuning stage.

\subsection{End-to-End Fine-Tunning for Task-Specific Optimization}\label{subsec: e2e_finetune}

Leveraging human priors eliminates the need for expensive multi-sensory data but introduces bias from human priors. Therefore, in the fine-tuning process, we train the model with a weakly supervised method on small-scale multi-modality data. We use 3,000 multi-sensory samples from both indoor and outdoor environments, including structured datasets \cite{neuhold2017mapillary, jiang2021rellis} and unstructured outdoor data \cite{jiang2021rellis}, all featuring 2D images paired with synchronized depth maps as inputs.

% \begin{figure}[htb]
% \vspace{-0.4cm}
%   \begin{center}
%     \includegraphics[width=0.3\textwidth]{Figures/PseudoLabel.pdf}
%   \end{center}
%   \captionsetup{font={small}}
%   \caption{Router label calculation.}
%     \label{fig:plabel}
%     \vspace{-0.4cm}
% \end{figure}
% As illustrated in Fig. \ref{fig:plabel}, given the set of expert traversability predictions ${\mathcal{T}_k}$ and the ground-truth traversability map $\mathcal{T}_{GT}$, we generate a routing supervision label $\mathcal{W}_{\text{sup}}$ by selecting, at each pixel, the prediction from the expert whose output most closely matches the ground truth.
Given the set of expert traversability predictions ${\mathcal{T}_k}$ and the ground-truth traversability map $\mathcal{T}_{GT}$, we generate a routing supervision label $\mathcal{W}_{\text{sup}}$ by selecting, at each pixel, the prediction from the expert whose output most closely matches the ground truth.
Specifically, for each expert $k$, the pixel-wise squared error at location $(i,j)$ is computed as $\varepsilon_k(i,j) = \left(\mathcal{T}_k(i,j) - \mathcal{T}_{GT}(i,j)\right)^2$. 
The routing supervising label at pixel $(i,j)$ is then determined by selecting the expert with the minimum error:
% \vspace{-1mm}
\begin{equation}
\small
    \mathcal{W}_{\text{sup}}(i,j) = \mathcal{T}_{k^*}(i,j), \quad \text{where} \quad k^* = \mathop{\arg\min}_{k \in \{1,2,\dots,K\}} \varepsilon_k(i,j)
% \vspace{-1mm}
\end{equation}

Applying this procedure across all pixel locations constructs the full supervision map for training the router.
Guided by the router label $\mathcal{W}_{\text{sup}}$, we unfreeze and jointly fine-tune both the routers and experts to promote better alignment and co-adaptation.
The domain router is trained to prioritize the expert whose prediction best matches the ground truth at each input, while each domain expert is fine-tuned to improve its performance on the terrain regions where it is selected as the best predictor. 
% Further details about the datasets are provided in \cref{sec:app_dataset}.

% we query all experts and compare their results with the ground-truth traversability map $\mathcal{T}_{GT}$ to calculate the pseudo labels. Specifically, after getting all $\{\mathcal{T}_k|k=1,2,...,K\}$ from all $K$ experts, we calculate the MSE error between each $\mathcal{T}_k$ with $\mathcal{T}_{GT}$: $\varepsilon_k = MSE(\mathcal{T}_k, \mathcal{T}_{GT})$. For each pixel, we choose the expert with least error as the ground truth of the pseudo label.

% Motivation: module co-optimize or co-adaptation

\section{Experiment Results}
\label{sec:exp}
\begin{figure*}[t!]
    \centering
        \includegraphics[width=0.95\linewidth]{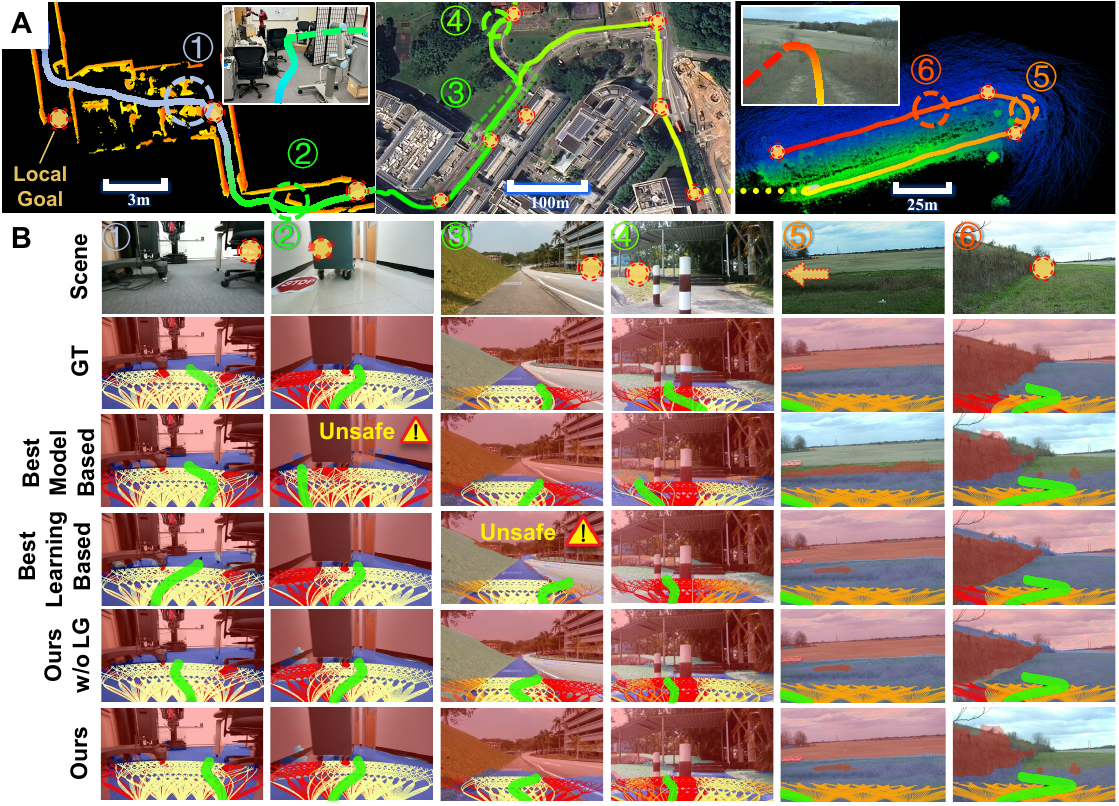}
        % \begin{tikzpicture}[remember picture, overlay, squarednode/.style={rectangle, opacity=.9, draw=black, fill=red!10, thick, minimum size=2mm},
        % title/.style={rectangle, fill=white, minimum size=2mm}
        % ]
        %     \node[squarednode, fill=white, inner sep=1.5pt] at (-6.62, 10.56) {A)};
        %     \node at (-6.68, 7.9) {B)};
        %     \node[title, rotate=90] at (-6.31, 7.4) {\scriptsize{\textbf{Scene}}};
        %     \node[title, rotate=90] at (-6.31, 6.3) {\scriptsize{\textbf{GT}}};
        %     \node[title,  rotate=90] at (-6.45, 5.1) {\scriptsize{\textbf{Model}}};
        %     \node[title,  rotate=90] at (-6.15, 5.1) {\scriptsize{\textbf{Based}}};
        %     \node[title,  rotate=90] at (-6.45, 3.85) {\scriptsize{\textbf{Learning}}};
        %     \node[title,  rotate=90] at (-6.15, 3.85) {\scriptsize{\textbf{Expert}}};
        %     \node[title,  rotate=90] at (-6.45, 2.65) {\scriptsize{\textbf{Ours}}};
        %     \node[title,  rotate=90] at (-6.15, 2.65) {\scriptsize{\textbf{w/o LG}}};
        %     \node[title,  rotate=90] at (-6.31, 1.35) {\scriptsize{\textbf{Ours}}};
        %     \node[title] at (1.56, 0.6) {\tiny{\textbf{Mixture Mode}}};
        %     \node[title] at (4.12, 0.6) {\tiny{\textbf{Transition Mode}}};
        %     \node[title] at (6.34, 0.6) {\tiny{\textbf{Unsafe Path}}};
        % \end{tikzpicture}
        \vspace{-0.2cm}
    \captionsetup{font={small}}
    \caption{\textbf{Sample frames from three domains.} 
    \textbf{(A)} Overview of the test scenes.
    \textbf{(B)} Representative frame with results: a semi-transparent traversability map is overlaid on the RGB image, red-to-blue means non-to-fully traversable; candidate motion primitives are drawn on the map and colored by their average traversability from red to gold; the green curve marks the final path. 
    % \textbf{(A) }Overview of the experiment scenes. \textbf{(B) } Visualization of the traversability map, motion primitives and final path. Traversability maps are overlayed on RGB images with transparency, red-blue represents non-to-fully traversable. Motion primitives are on top of each image, their traversability from non-to-full are marked as red-yellow-gold. Green curves represent the final path.
    % We qualitatively evaluate the traversability map projection on query frames for Model-based,  Learning experts, and our approach over different domains. 
    %  Frames 1 and 2 are from our indoor dataset, frames 3 and 4 are from a trajectory in an outdoor structured environment \cite{zhang2023cntu}, and frames 5 and 6 are from an outdoor unstructured environment \cite{jiang2021rellis}.
    }
    \label{fig:exp}
    \vspace{-0.7cm}
\end{figure*}

To evaluate the cross-domain traversability estimation ability and computational efficiency of \AlgName{}, we compare our approach with 4 state-of-the-art baselines that can run in real-time: Falco~\cite{zhang2020falco}, Mask2Former~\cite{cheng2021mask2former} for indoor, MaskFormer~\cite{cheng2021maskformer} for structured outdoor, and GANav~\cite{guan2021ganav} for unstructured outdoor environment. Additionally, we include four ablation studies, \AlgName{} without fine-tuning and lazy gating (Ours w/o FT/LG), without lazy gating (Ours w/o LG), without lower-level lazy gating (Ours w/o LLG) and without fine-tuning (Ours w/o FT), to evaluate the individual contributions of these components to overall performance.
Scene overviews for all three domains are shown in \cref{fig:exp}.A. The indoor experiment is conducted in the real world, while outdoor experiments use test sequences from \cite{jiang2021rellis, zhang2023cntu}.
% To thoroughly assess the generalization capabilities of each model, we evaluate them across a diverse range of environments. This includes indoor scenarios from a dataset we collected, as well as unstructured and structured outdoor environments, using data from \cite{jiang2021rellis, zhang2023cntu}.

We evaluate performance using four metrics: \textbf{Motion Primitives Quality: } We introduce aligned-path traversability estimation mean square error ($e_P$) to measure the mean squared error between estimated $\mathcal{T}$ and ground-truth traversability $\mathcal{T}_\textrm{gt}$ along candidate motion primitives, which directly reflects planning quality with arbitrary goals. It can be formulated as: $e_P = \left\lVert S(\mathcal{T}) - S(\mathcal{T}_\textrm{gt}) \right\rVert_2^2 \big/ \left| S(\mathcal{T}_\textrm{gt}) \right|$. 
\textbf{Traversability Map Quality: } We introduce traversability map estimation mean square error ($e_M$) to assess the accuracy of the predicted traversability map; 
\textbf{Computational Efficiency: } is quantified in GFLOPS; and 
\textbf{Normalized path quality ($Q_p$):} Given any path $\tau$, goal $p_\textrm{goal}$, and $\mathcal{T}$, we define the path quality as $\eta(\tau) = 1 - f(\mathcal{T}, p_{\mathrm{goal}})$, where $f(\mathcal{T}, p_{\mathrm{goal}}) \in [0,1]$ is defined in \cref{eq:planner}). The path quality ratio $Q_{p}$ is then formulated as: $Q_{p} = \eta(\hat{\tau}) / \eta(\tau_\textrm{gt})$, where $\tau_\textrm{gt}$ represents the optimal ground truth path and $\hat{\tau}$ means the estimated path.  This metric ranges between 0 and 1, with higher values indicating better path quality.
% (4) Normalized path quality ($Q_p$) compares the reward of the predicted path to the GT one, indicating the effectiveness of the generated map for navigation.
Since our approach involves the lazy gating mechanism that doesn't need to produce a full $\mathcal{T}$ for path planning, we do not record $e_P$ and $e_M$ for our approach.

We report GFLOPs instead
of runtime because some experts in our MoE framework are implemented in C++
while others in Python. Thus, we believe GFLOPs provide a fairer metric for
comparison. We here
report the average runtimes estimated in outdoor scenario to demonstrate our time efficiency: Gating network 6.0ms;
Model-based expert 6.2ms; Learning-based expert 64.3ms; and path calculation
119.8ms, and the average processing time is 163.8ms. We believe this is sufficient for
common navigation tasks, although the code implementation is not optimized yet.

% \textbf{Motion Primitives Quality: } 
% The ultimate goal of generating traversability maps in navigation is to optimize the trajectory. Therefore, we utilize Path-MSE directly measure the quality of the traversability estimation for all candidate motion primitives. The formulation is:\\
% \begin{center}
%     \vspace{-0.5cm}
%     $P_{mse} = \left\lVert S(\mathcal{T}) - S(\mathcal{T}_{gt}) \right\rVert_2^2 \big/ \left| S(\mathcal{T}_{gt}) \right|$
% \end{center}

% \textbf{Traversability Map Quality:} We adopt image-based MSE to evaluate the quality of $\mathcal{T}$ compared to $\mathcal{T}_{gt}$. Though not directly related to the navigation performance, $I_{mse}$ measures general environmental perception ability, which is important for other related tasks.

% \textbf{Computational Efficiency: }We quantify the required floating-point operations during inference using GFLOPS (Giga Floating-Point Operations per Second). 

% \textbf{Path Quality:} Given any path $\tau$, goal $p_{goal}$, and $\mathcal{T}$, we define the path quality as $\eta(\tau) = 1 - f(\tau;\mathcal{T}_{gt}, p_{goal})$, where $f(\tau;\mathcal{T}, p_{goal}) \in [0,1]$ denotes the path cost (as defined in \cref{eq:planner}). Path quality ratio $Q_{p}$ is then formulated as: $Q_{p} = \eta(\tau_{est}) / \eta(\tau_{gt})$. This metric ranges between 0 and 1, with higher values indicating better path quality.

\noindent\textbf{Quantitative Results.\quad} 
As illustrated in \cref{fig:exp}\textcolor{anotBlue}{.B}, our method navigates complex environments, such as \cref{fig:exp}\textcolor{anotBlue}{.B}\textccir{1}, safely by primarily leveraging model-based experts while querying learning-based experts only when semantic cues are critical, achieving robust performance across all domains and terrain types. In \cref{fig:exp}\textcolor{anotBlue}{.B}\textccir{3}, learning-based expert \cite{cheng2021maskformer} fail to strictly adhere to safety constraints, potentially causing the robot falling off a curb into traffic, while \AlgName{} utilizes geometric perception from model-based methods to maintain safety. Conversely, in \cref{fig:exp}\textcolor{anotBlue}{.B}\textccir{2} and \textccir{4}, the learning-based experts' semantic perception correctly identifies non- or low-traversable areas (e.g., stop sign, grass), whereas the model-based method \cite{zhang2020falco} mistakenly classify grass as traversable due to its geometric similarity to planar surfaces.

\noindent\textbf{Cross-Domain Traversability Estimation Accuracy.\quad} 
As shown in \cref{tab:performance}, \AlgName{} is the only method that can generalize well across all three domains, achieving best or second best performance among $e_P$, $e_M$ and $Q_p$. 
In structured outdoor environment where learning-based expert \cite{cheng2021maskformer} and Model-based one \cite{zhang2020falco} have complementary expertise, the advantage of \AlgName{} becomes significant and we achieve the best performance among $e_P$, $e_M$ and $Q_p$. 
In indoor environment, our approach achieves comparable accuracy compared to domain-specific baseline \cite{cheng2021mask2former} with only a $0.03$ drop in $Q_p$. The same phenomenon extends to unstructured outdoor environment with a $0.01$ drop in $Q_p$. 
This trend aligns with findings reported in other MoE works \cite{wang2023fusing, cai2024survey}, which indicates that MoE methods tend to deliver similar or occasionally slightly lower performance in domains where a single expert already excels. Additionally, our results are impacted by hardware misalignment and synchronization issues between the RGB camera and lidar or depth camera, which can be observed in the model-based method part in \cref{fig:exp}.
\textbf{Effectiveness of FT.\quad }
Comparing \AlgName{} w/o FT/LG to \AlgName{} w/o LG, fine-tuning improves estimation accuracy up to $41\%$ in $e_P$ and $54\%$ in $e_M$. When lazy gating is applied (Ours w/o FT v.s. Ours in \cref{tab:performance}), fine-tuning balances $Q_p$ and GFLOPS. Specifically, in the indoor scenario, fine-tuning shifts weights toward model-based methods, improving efficiency with small $Q_p$ loss. Conversely, in outdoor environments, fine-tuning allocates more weights to learning-based methods, enhancing $Q_p$.

\noindent\textbf{Efficiency and Effectiveness for Path Planning.\quad} 
Across all domains, \AlgName{} has much higher efficiency than the best performing expert, marked in blue in \cref{tab:performance}, by up to $76\%$. With the guaranteed solution quality introduced by the lazy gating mechanism, 
% Based on the planning horizon, the bound $\delta$ is set as $0.05, 0.1, 0.2$ for indoor, outdoor structured and outdoor unstrctured domains, 
the resulting $Q_p$ are within $0.03$ compared to \AlgName{} without LG, which is within the bound set as $0.05$.

\noindent\textbf{Effectiveness of LG.\quad}
Comparing \AlgName{} to our method without LG, lazy gating can decrease the computational overhead up to $96.3\%$, with only up to $0.02$ loss in $Q_p$. Interestingly, we found the upper level lazy router achieves over $95\%$ confidence and over $98\%$ in accuracy across all domains, meaning that keeping the domain lazy router will not affect the estimation accuracy but significantly improve the computational efficiency, shown in \cref{tab:performance}: Ours w/o LG and Ours w/o LLG. In this case, our method can still improve the computational efficiency by up to $82.6\%$, thereby reducing the average inference time to 164 milliseconds.

Overall, domain-specific approaches show satisfying performance within their area of expertise, they exhibit significantly weaker performance in out-of-distribution scenarios. \AlgName{} demonstrates more robust and effective corss-domain traversability estimation accuracy.
This reinforces our intuition that a robust and effective navigation model can be achieved by hierarchically combining experts with different expertise, paired with an efficient routing mechanism.
Furthermore, our domain-gating mechanism allows the model to seamlessly scale with new learning-based navigation models across any domain, requiring only minimal fine-tuning of the gating networks.

% \begin{wraptable}{c}{\textwidth}
\begin{table*}[h!]
\resizebox{\textwidth}{!}{
\large
\begin{tabular}{c|cccc|cccc|cccc?cccc}
\toprule
\multirow{3}{*}{\textbf{Methods}}
  & \multicolumn{12}{c}{\textbf{Environments}} \\ 
  & \multicolumn{4}{c|}{\textit{Indoor}}
  & \multicolumn{4}{c|}{\textit{Out. Struc.}}
  & \multicolumn{4}{c?}{\textit{Out. Unstruc.}}
  & \multicolumn{4}{c}{\textit{Avg}}
  \\ 
  & $e_P$ $\downarrow$ & $e_M$ $\downarrow$ & $Q_{p}$ $\uparrow$ & GFLOPS $\downarrow$
  & $e_P$ $\downarrow$ & $e_M$ $\downarrow$ & $Q_{p}$ $\uparrow$ & GFLOPS $\downarrow$
  & $e_P$ $\downarrow$ & $e_M$ $\downarrow$ & $Q_{p}$ $\uparrow$ & GFLOPS $\downarrow$ 
  & $e_P$ $\downarrow$ & $e_M$ $\downarrow$ & $Q_{p}$ $\uparrow$ & GFLOPS $\downarrow$ \\
\midrule
Falco~\cite{zhang2020falco}  & 0.213 & 0.100 & 0.87 & 0.13 & 0.157 & 0.098 & 0.9 & 0.13 & 0.104 & 0.048 & 0.64 & 0.13 & 0.158 & 0.082 & 0.74 & \textbf{0.13}\\
Mask2Former~\cite{cheng2021maskformer}   & \cellcolor{myblue!50} \textbf{0.021} & \cellcolor{myblue!50} \textbf{0.012} & \cellcolor{myblue!50}\textbf{0.98} & \cellcolor{myblue!50} 246.31 & 0.091 & 0.033 & 0.87 & 246.31 & 0.083 & 0.057  & 0.80 & 246.31 & 0.065 & 0.034 & 0.88 & 246.31 \\
MaskFormer\cite {cheng2021mask2former} & 0.470 & 0.187 & 0.74 & 48.3 & \cellcolor{myblue!50} 0.070 & \cellcolor{myblue!50}\underline{0.026} & \cellcolor{myblue!50} 0.92 & \cellcolor{myblue!50} 48.3 & 0.069 & \underline{0.033} & 0.81 & 48.3 & 0.203 & 0.082 & 0.83 & 48.3 \\
GaNav~\cite{guan2021ganav}                 & 0.811 & 0.403 & 0.33 & 18.69 & 0.729 & 0.265 & 0.17 & 18.69 & \cellcolor{myblue!50}\textbf{0.034} & \cellcolor{myblue!50}\textbf{0.031} & \cellcolor{myblue!50}\textbf{0.87} & \cellcolor{myblue!50} 18.69 & 0.525 & 0.233 & 0.56 & 18.69 \\
\midrule
Ours w/o FT/LG      & 0.039 & 0.028 & \textbf{0.98} & 313.18 & \underline{0.064} & 0.034 & 0.96 & 313.18 & 0.048 & 0.035 & 0.84 & 313.18 & \underline{0.051} & \underline{0.032} & 0.92 & 313.18 \\
Ours w/o LG     & \underline{0.023} & \underline{0.013} & \underline{0.97} & 313.18 & \textbf{0.055} & \textbf{0.021} & \textbf{0.98} & 313.18 & \underline{0.043} & \underline{0.033} & \textbf{0.87} & 313.18 & \textbf{0.040} & \textbf{0.022} & \textbf{0.94} & 313.18 \\
Ours w/o LLG     & \underline{0.023} & \underline{0.013} & \underline{0.97} & 252.58 & \textbf{0.055} & \textbf{0.021} & \textbf{0.98} & 54.57 & \underline{0.043} & \underline{0.033} & \textbf{0.87} & 24.96 & \textbf{0.040} & \textbf{0.022} & \textbf{0.94} & 110.7 \\
Ours w/o FT    & - & - & 0.95 & 122.04 & - & - & 0.96    & 9.47 & - & - & 0.84 & 10.28 & - & - & 0.92 & 52.16 \\
\rowcolor{orange!30} Ours     & - & - & 0.95 & 94.83 & - & - & \textbf{0.98} & 11.58 & - & - & \underline{0.86} & 16.25 & - & - & \underline{0.93} & \underline{40.22} \\
\bottomrule
\end{tabular}
}
\captionsetup{font={small}}
\caption{ \textbf{Quantitative results} of traversability estimation accuracy, path quality, and computational efficiency over multiple datasets.}
\label{tab:performance}
\vspace{-0.7cm}

% \end{wraptable}
\end{table*}

\section{Conclusion} 
% \vspace{-0.3cm}
We presented \AlgName{}, a hierarchical and modular approach that combines the fast, reliable model-based methods with the semantic reasoning of learning-based approaches. Our modular design is easily extendable with new experts, allowing better adaptation to diverse environments. By developing a hierarchical Mixture of Experts (MoE) architecture, we adaptively leverage specialized experts for different terrains, enhancing generalization across both seen and unseen environments.
Key innovations include a lazy gating mechanism for improved computational efficiency and a two-stage training strategy that reduces reliance on high-quality multi-sensory data. Extensive experiments show that \AlgName{} reduces computational cost by 81.2\% while maintaining path planning quality, outperforming existing methods in both efficiency and generalization beyond training domain.
Our approach is highly extensible, enabling future breakthroughs in robotics by unifying non-differentiable and differentiable approaches across a variety of tasks.
\label{sec:conclusion}

\bibliographystyle{IEEEtran}
\bibliography{references}

% \appendix
% \include{appendix}

\end{document}